\documentclass[10pt,twocolumn,letterpaper]{article}

\usepackage[pagenumbers]{cvpr} 

\usepackage{graphicx}
\usepackage{subcaption}
\usepackage{amsmath}
\usepackage{amssymb}
\usepackage{booktabs}
\usepackage{multirow}

\usepackage[pagebackref,breaklinks,colorlinks]{hyperref}

\usepackage[capitalize]{cleveref}
\crefname{section}{Sec.}{Secs.}
\Crefname{section}{Section}{Sections}
\Crefname{table}{Table}{Tables}
\crefname{table}{Tab.}{Tabs.}

\def\cvprPaperID{*****} 
\def\confName{CVPR}
\def\confYear{2022}

\begin{document}

\title{PP-LiteSeg: A Superior Real-Time Semantic Segmentation Model}

\author{Juncai Peng, Yi Liu, Shiyu Tang, Yuying Hao, Lutao Chu,  Guowei Chen,\\ Zewu Wu, Zeyu Chen, Zhiliang Yu, Yuning Du, Qingqing Dang, \\ Baohua Lai, Qiwen Liu, Xiaoguang Hu, Dianhai Yu, Yanjun Ma\\
        Baidu Inc.\\
{\tt\small \{pengjuncai, liuyi22\}@baidu.com}
}
\maketitle

\begin{abstract}
Real-world applications have high demands for semantic segmentation methods.
Although semantic segmentation has made remarkable leap-forwards with deep learning, the performance of real-time methods is not satisfactory.
In this work, we propose PP-LiteSeg, a novel lightweight model for the real-time semantic segmentation task.
Specifically, we present a Flexible and Lightweight Decoder (FLD) to reduce computation overhead of previous decoder.
To strengthen feature representations, we propose a Unified Attention Fusion Module (UAFM), which takes advantage of spatial and channel attention to produce a weight and then fuses the input features with the weight.
Moreover, a Simple Pyramid Pooling Module (SPPM) is proposed to aggregate global context with low computation cost.
Extensive evaluations demonstrate that PP-LiteSeg achieves a superior trade-off between accuracy and speed compared to other methods.
On the Cityscapes test set, PP-LiteSeg achieves 72.0\% mIoU/273.6 FPS and 77.5\% mIoU/102.6 FPS on NVIDIA GTX 1080Ti.
Source code and models are available at PaddleSeg: \href{https://github.com/PaddlePaddle/PaddleSeg}{https://github.com/PaddlePaddle/PaddleSeg}.

\end{abstract}

\begin{figure}[t]
    \centering
    \includegraphics[width=\linewidth]{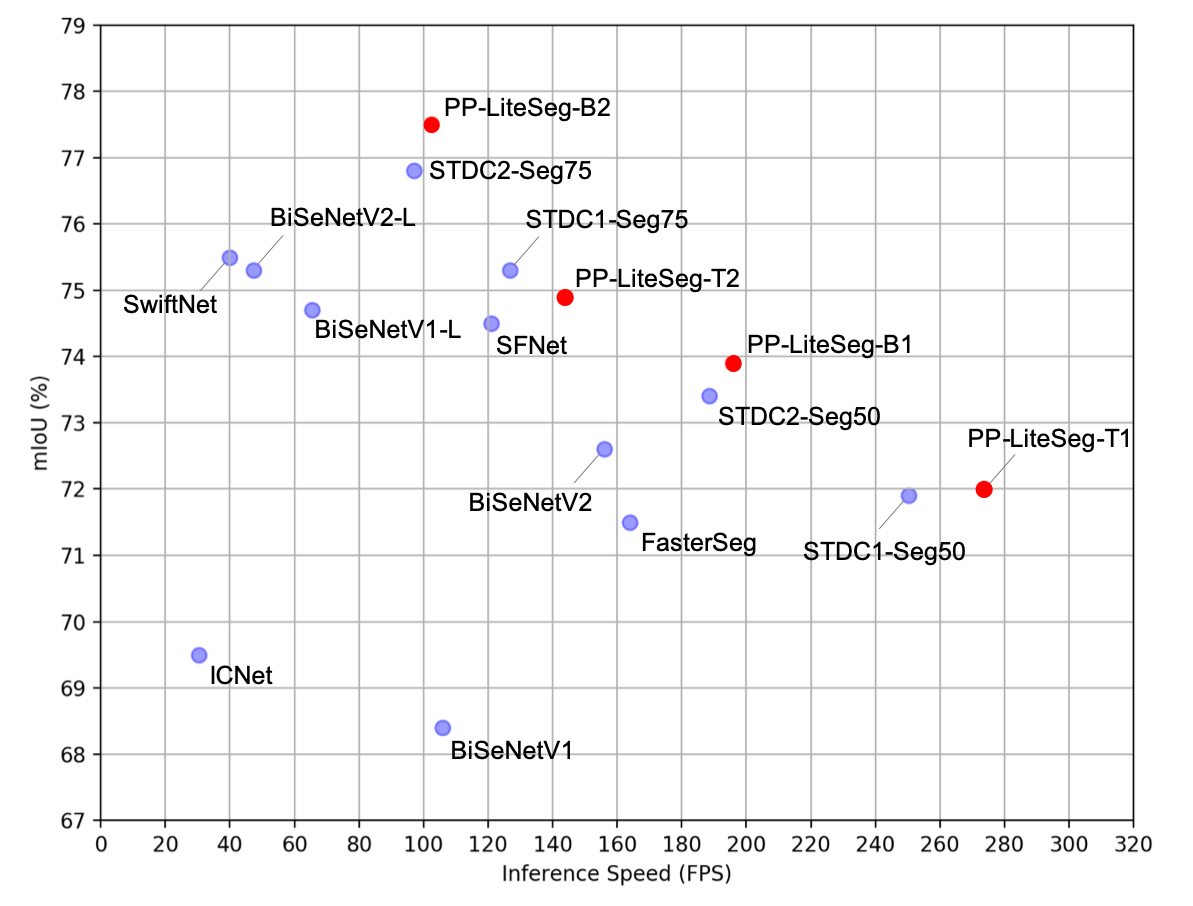}
    \caption{The comparison of segmentation accuracy (mIoU) and inference speed (FPS) on the Cityscapes test set. The red dots represent our proposed PP-LiteSeg. The testing device is NVIDIA GTX 1080Ti. The experimental results demonstrate that PP-LiteSeg achieves state-of-the-art trade-off between accuracy and speed. }
    \label{fig:speed_iou_fig}
\end{figure}

\section{Introduction}
\label{sec:introdution}


Semantic segmentation aims to precisely predict the label of each pixel in an image.
It has been widely applied in real-world applications, e.g. medical imaging \cite{zhou2018unet++}, autonomous driving \cite{siam2018comparative, hou2019learning}, video conferencing \cite{chu2022pp}, semi-automatic annotation~\cite{hao2021edgeflow}. As a fundamental task in computer vision, semantic segmentation has attracted a lot of attention from researchers \cite{liu2021paddleseg, lateef2019survey}.

With the remarkable progress of deep learning, a lot of semantic segmentation methods have been proposed based on convolutional neural network \cite{long2015fully, zhao2017pyramid, chen2018encoder, li2020semantic, yu2021bisenet}. 
FCN \cite{long2015fully} is the first fully convolutional network trained in an end-to-end and pixel-to-pixel way.
It also presents the primitively encoder-decoder architecture in semantic segmentation, which is widely adopted in subsequent methods.
To achieve higher accuracy, PSPNet \cite{zhao2017pyramid} utilizes a pyramid pooling module to aggregate global context and SFNet \cite{li2020semantic} proposes a flow alignment module to strengthen the feature representations.

Yet, these models are not suitable for real-time applications because of their high computation cost. To accelerate the inference speed, Espnetv2\cite{mehta2019espnetv2} utilizes light-weight convolutions to extract features from an enlarged receptive field.
BiSeNetV2 \cite{yu2021bisenet} proposes bilateral segmentation network and extracts the detail features and semantic features separately.
STDCSeg \cite{fan2021rethinking} designs a new backbone named STDC to improve the computation efficiency.
However, these models do not achieve satisfactory trade-off between accuracy and speed.

\begin{figure*}[t]
  \centering
   \includegraphics[width=0.7\linewidth]{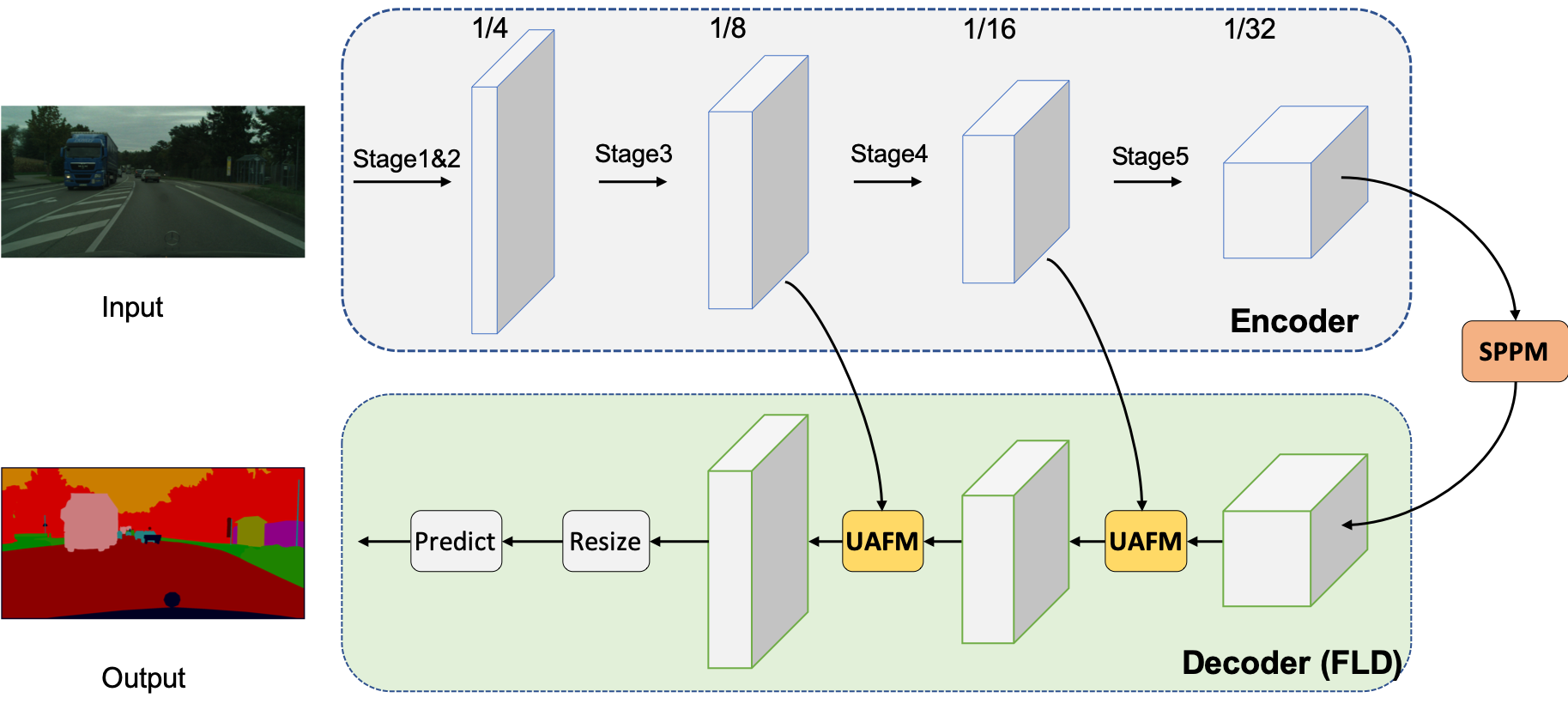}
   \caption{The architecture overview. PP-LiteSeg consists of three modules: encoder, aggregation and decoder. A lightweight network is used as encoder to extract the features from different levels. The Simple Pyramid Pooling Module (SPPM) is responsible for aggregating the global context. The Flexible and Lightweight Decoder (FLD) fuses the detail and semantic features from high level to low level and outputs the result. Remarkably, FLD uses the Unified Attention Fusion Module (UAFM) to strengthen feature representations. }
   \label{fig:arch}
\end{figure*}

In this work, we propose a real-time and hand-craft network named PP-LiteSeg. As illustrated in Figure \ref{fig:arch}, PP-LiteSeg adopts the encode-decoder architecture and consists of three novel modules: Flexible and Lightweight Decoder (FLD), Unified Attention Fusion Module (UAFM) and Simple Pyramid Pooling Module (SPPM). The motivations and details of these modules are introduced as follows.

The encoder in semantic segmentation models extracts hierarchical features, and the decoder fuses and unsamples features.
For the features from low level to high level in encoder, the number of channels increases and the spatial size decreases, which is an efficient design.
For the features from high level to low level in decoder, the spatial size increases, while the number of channels are the same in recent models \cite{li2020semantic, fan2021rethinking}.
Therefore, we present a Flexible and Lightweight Decoder (FLD), which gradually reduces the channels and increases the spatial size for the features.
Besides, the volume of proposed decoder can be easily adjusted according to the encoder. The flexible design balances the computation complexity of encoder and decoder, which makes the overall model more efficient.


Strengthening feature representations is a crucial way to improve segmentation accuracy \cite{li2020semantic, huang2021fapn, song2021attanet}.
It is usually achieved through fusing the low-level and high-level features in a decoder. However, the fusion modules in existing methods usually suffer from high computation cost.
In this work, we propose a Unified Attention Fusion Module (UAFM) to strengthen feature representations efficiently. As shown in Figure \ref{fig:uafm}, UAFM first takes advantage of the attention module to produce weight $\alpha$, and then fuses the input features with $\alpha$.
In UAFM, there are two kinds of attention modules, i.e. spatial and channel attention modules, which exploit inter-spatial and inter-channel relationships of the input features.

Contextual aggregation is another key to promote segmentation accuracy, but previous aggregation modules are time-consuming for real-time networks.
Based on the framework of PPM \cite{zhao2017pyramid}, we design a Simple Pyramid Pooling Module (SPPM), which reduces the intermediate and output channels, removes the short-cut, and replaces the concatenate operation with an add operation. Experimental results show SPPM contributes to the segmentation accuracy with low computation cost.

We evaluate the proposed PP-LiteSeg through extensive experiments on Cityscapes and CamVid dataset. As illustrated in Figure \ref{fig:speed_iou_fig}, PP-LiteSeg achieves a superior trade-off between segmentation accuracy and inference speed. Specifically, PP-LiteSeg achieves 72.0\% mIoU/273.6 FPS and 77.5\% mIoU/102.6 FPS on the Cityscapes test set.

Our main contributions are summarized as follows:
\begin{itemize}
\item We propose a Flexible and Lightweight Decoder (FLD), which mitigates the redundancy of the decoder and balances the computation cost of the encoder and decoder.
\item We present a Unified Attention Fusion Module (UAFM) that utilizes channel and spatial attention to strengthen the feature representations.
\item We propose a Simple Pyramid Pooling Module (SPPM) to aggregate global context. SPPM promotes the segmentation accuracy with minor extra inference time.
\item Based on the above modules, we propose PP-LiteSeg, a real-time semantic segmentation model. Extensive experiments demonstrate our SOTA performance in terms of accuracy and speed.
\end{itemize}

\section{Related Work}
\label{sec:related_work}

\subsection{Semantic Segmentation}
Deep learning has helped semantic segmentation make remarkable leap-forwards.
FCN \cite{long2015fully} is the first full convolutional network for semantic segmentation. It is trained in an end-to-end and pixel-to-pixel way. Besides, images with arbitrary size can be segmented by FCN.
Following the design of FCN, various methods have been proposed in later. 
Segnet\cite{badrinarayanan2017segnet} applies the indices of max-pooling operation in encoder to upsampling operation in decoder. Therefore, the information in decoder is reused and the decoder produces refined features.
PSPNet\cite{zhao2017pyramid} proposes the pyramid pooling module to aggregate local and global information, which is effective for segmentation accuracy.
Besides, recent semantic segmentation methods \cite{huang2021fapn, liu2021swin} utilize the transformer architecture to achieve better accuracy.

\subsection{Real-time Semantic Segmentation}

To fulfill the real-time demands of semantic segmentation, lots of methods have been proposed, e.g lightweight module design \cite{fan2021rethinking, mehta2019espnetv2}, dual-branch architecture \cite{yu2018bisenet, yu2021bisenet}, early-downsampling strategy \cite{paszke2016enet}, multiscale image cascade network \cite{zhao2018icnet}.
ENet \cite{paszke2016enet} uses an early-downsampling strategy to reduce the computation cost of processing large images and feature maps.
For efficiency, ICNet \cite{zhao2018icnet} designs a multi-resolution image cascade network.
Based on bilateral segmentation network, Bisenet \cite{yu2021bisenet} extracts the detail features and semantic features separately. The bilateral network is lightweight, so the inference speed is fast.
STDCSeg \cite{fan2021rethinking} proposes the channel-reduced receptive field-enlarged STDC module and designs an efficient backbone, which can strengthen the feature representations with low computation cost. To eliminate the redundancy in two-branch network, STDCSeg guides the features with detailed ground truth, so the efficiency is further improved.
Espnetv2\cite{mehta2019espnetv2} uses group point-wise and depth-wise dilated separable convolutions to learn features from an enlarged receptive field in a computation friendly manner.

\subsection{Feature Fusion Module}
The feature fusion module is commonly used in semantic segmentation to strengthen feature representations.
In addition to the element-wise summation and concatenation methods, researchers propose several methods as follows.
In BiSeNet \cite{yu2021bisenet}, the BGA module employs element-wise mul method to fuse the features from the spatial and contextual branches.
To enhance the features with high-level context, DFANet \cite{li2019dfanet} fuses features in a stage-wise and subnet-wise way.
To tackle the problem of misalignment, SFNet \cite{li2020semantic} and AlignSeg \cite{huang2021alignseg} first learn the transformation offsets through a CNN module, and then apply the transformation offsets to grid\_sample operation to generate the refined feature.
In detail, SFNet designs the flow alignment module.
AlignSeg designs aligned feature aggregation module and the aligned context modeling module.
FaPN \cite{huang2021fapn} solves the feature misalignment problem by applying the transformation offsets to deformable convolution.

\section{Proposed Method}
\label{sec:proposed_method}

In this section, we first introduce Flexible and Lightweight Decoder (FLD), Unified Attention Fusion Module (UAFM) and Simple Pyramid Pooling Module (SPPM), respectively. Then, we present the architecture of PP-LiteSeg for real-time semantic segmentation.

\subsection{Flexible and Lightweight Decoder}

\begin{figure}[t]
    \centering
    \includegraphics[width=1.0\linewidth]{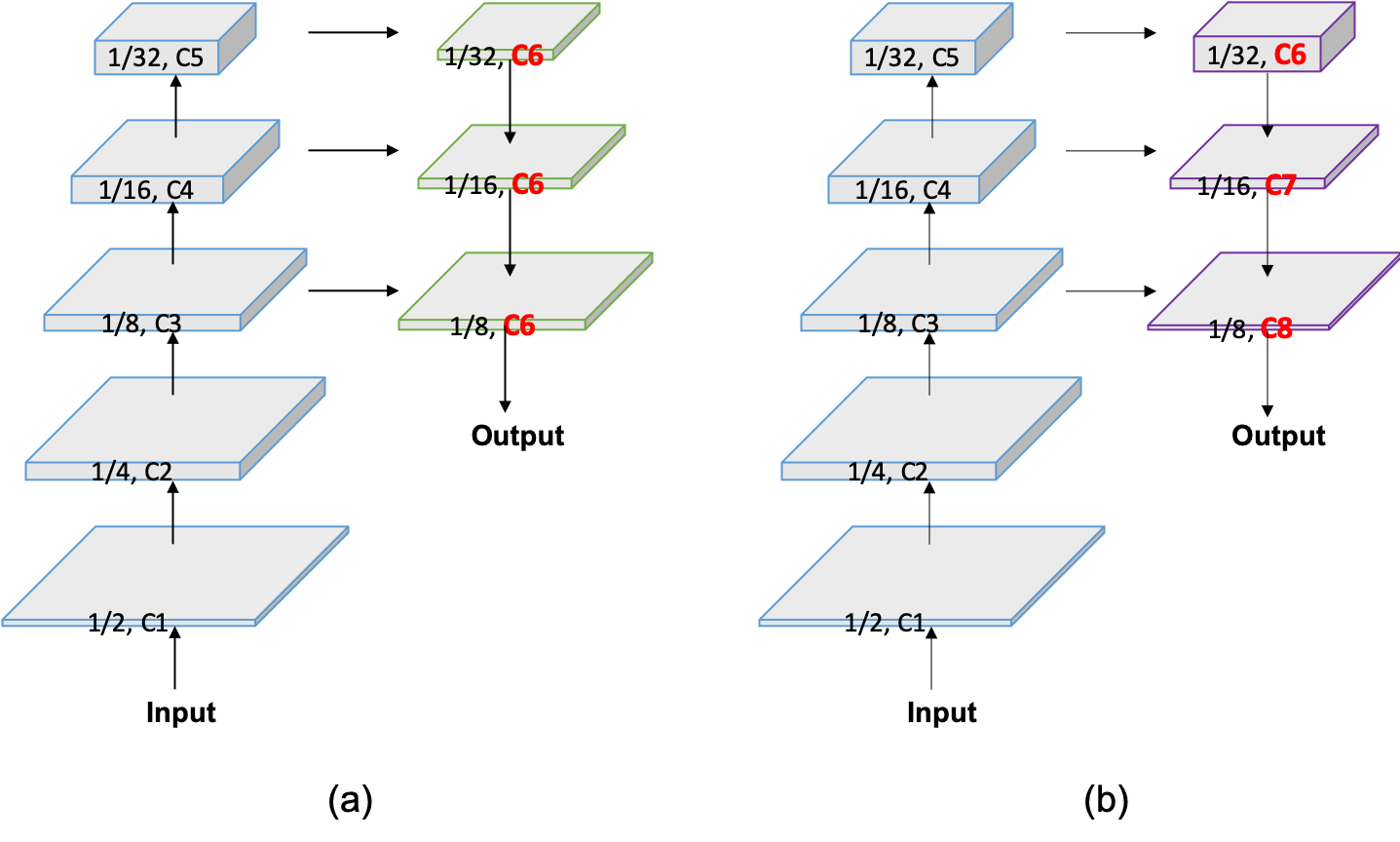}
    \caption{(a) The conventional encoder-decoder, in which the decoder keeps the features channels the same. (b) The encoder and proposed Flexible and Lightweight Decoder (FLD). FLD gradually reduces the channels for the features from high level to low level. Moreover, the volume of FLD is adjusted to conform to the encoder.}
    \label{fig:fld}
\end{figure}

Encoder-decoder architecture has been proved to be effective for semantic segmentation.
In general, the encoder utilizes a series of layers grouped into several stages to extract hierarchical features. For the features from low level to high level, the number of channels gradually increases and the spatial size of the features decreases. This design balances the computation cost of each stage, which ensures the efficiency of the encoder.
The decoder also has several stages, which are responsible for fusing and upsampling features.
Although the spatial size of features increases from high level to low level, the decoder in recent lightweight models keeps the feature channels the same in all levels. Therefore, the computation cost of shallow stage is much larger than that of the deep stage, which leads to the computation redundancy in shallow stage.
To improve the efficiency of decoder, we present a Flexible and Lightweight Decoder (FLD). 
As illustrated in Figure \ref{fig:fld}, FLD gradually decreases the channels of the features from high level to low level.
FLD can easily adjusts the computation cost to achieve better balance between encoder and decoder.
Although the channels of features in FLD are decreasing, our experiments show that PP-LiteSeg achieves competitive accuracy compared to other methods.

\subsection{Unified Attention Fusion Module}

As discussed above, fusing multi-level features is essential to achieve high segmentation accuracy.
In addition to the element-wise summation and concatenation methods, researchers propose several methods, e.g. SFNet \cite{li2020semantic}, FaPN \cite{huang2021fapn} and AttaNet \cite{song2021attanet}.
In this work, we propose a Unified Attention Fusion Module (UAFM) that applies channel and spatial attention to enrich the fused feature representations.

\begin{figure}[t]
  \centering
   \includegraphics[width=1.0\linewidth]{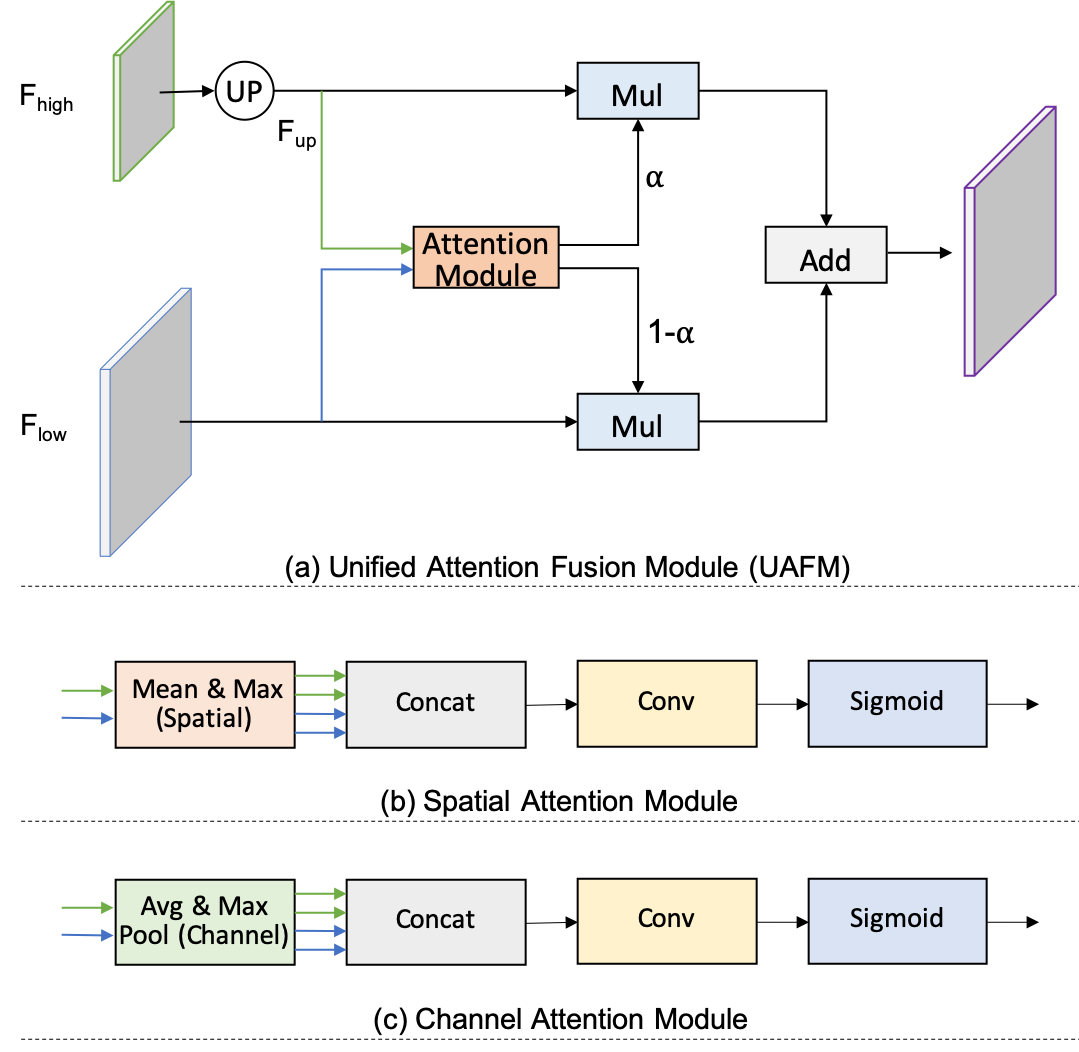}
   \caption{(a) The framework of Unified Attention Fusion Module (UAFM). (b) The Spatial Attention Module. (c) The Channel Attention Module. UAFM first uses spatial or channel attention modules to produce the weight $\alpha$, and then fuses the input features with Mul and Add operation.}
   \label{fig:uafm}
\end{figure}

\noindent
\textbf{UAFM Framework.} As illustrated in Figure \ref{fig:uafm} (a), UAFM utilizes an attention module to produce the weight $\alpha$, and fuses the input features with $\alpha$ by Mul and Add operations. 
In detail, the input features are denoted as $F_{high}$ and $F_{low}$.
$F_{high}$ is the output of the deeper module, and $F_{low}$ is the counterpart from the encoder. Note that they have same channels.
UAFM first makes use of bilinear interpolation operation to upsample $F_{high}$ to the same size of $F_{low}$, while the upsampled feature is denoted as $F_{up}$.
Then, the attention module takes $F_{up}$ and $F_{low}$ as input and produces the weight $\alpha$. Note that, the attention module can be a plugin, such as spatial attention module, channel attention module, etc.
After that, to obtain attention-weighted features, we apply the element-wise Mul operation to $F_{up}$ and $F_{low}$, respectively.
Finally, UAFM performs element-wise addition for the attention-weighted features and outputs the fused feature.
We can formulate the above procedure as equation \ref{eq:uafm}.

\begin{equation}
\begin{aligned}
F_{up} &= Upsample(F_{high}) \\
\alpha &= Attention(F_{up}, F_{low}) \\
F_{out} &= F_{up} \cdot \alpha + F_{low} \cdot (1 - \alpha)
\end{aligned}
\label{eq:uafm}
\end{equation}

\noindent
\textbf{Spatial Attention Module.}
The motivation of the spatial attention module is exploiting the inter-spatial relationship to produce a weight, which represents the importance of each pixel in the input features.
As shown in Figure \ref{fig:uafm} (b), given the input features, i.e., $F_{up} \in R^{C \times H \times W}$ and $F_{low} \in R^{C \times H \times W}$, we first perform mean and max operations along the channel axis to generates four features, of which the dimension is $R^{1 \times H \times W}$. Afterwards, these four features are concatenated to a feature $F_{cat} \in R^{4 \times H \times W}$. For the concatenated feature, the convolution and sigmoid operations are applied to output $\alpha \in R^{1 \times H \times W}$. The formulation of the spatial attention module is shown in equation \ref{eq:sam}.
Furthermore, the spatial attention module can be flexible, e.g. removing the max operation to reduce computation cost.

\begin{equation}
\begin{aligned}
F_{cat} &= Concat(Mean(F_{up}), Max(F_{up}), \\
        & Mean(F_{low}), Max(F_{low})) \\
\alpha &= Sigmoid(Conv(F_{cat}))
\end{aligned}
\label{eq:sam}
\end{equation}

\noindent
\textbf{Channel Attention Module.}
The key concept of the channel attention module is leveraging the inter-channel relationship to generate a weight, which indicates the importance of each channel in the input features.
As shown in Figure \ref{fig:uafm} (b), the proposed channel attention module utilizes average-pooling and max-pooling operations to squeeze the spatial dimension of input features. This procedure generates four features with the dimension $R^{C \times 1 \times 1}$.
Then, it concatenates these four features along the channel axis and performs convolution and sigmoid operations to produce a weight $\alpha \in R^{C \times 1 \times 1}$.
In short, the procedures of the channel attention module can be formulated as equation \ref{eq:cam}.

\begin{equation}
\begin{aligned}
F_{cat} &= Concat(AvgPool(F_{up}), MaxPool(F_{up}), \\
        & AvgPool(F_{low}), MaxPool(F_{low})) \\
\alpha &= Sigmoid(Conv(F_{cat}))
\end{aligned}
\label{eq:cam}
\end{equation}

\subsection{Simple Pyramid Pooling Module}

\begin{figure}[t]
  \centering
   \includegraphics[width=1.0\linewidth]{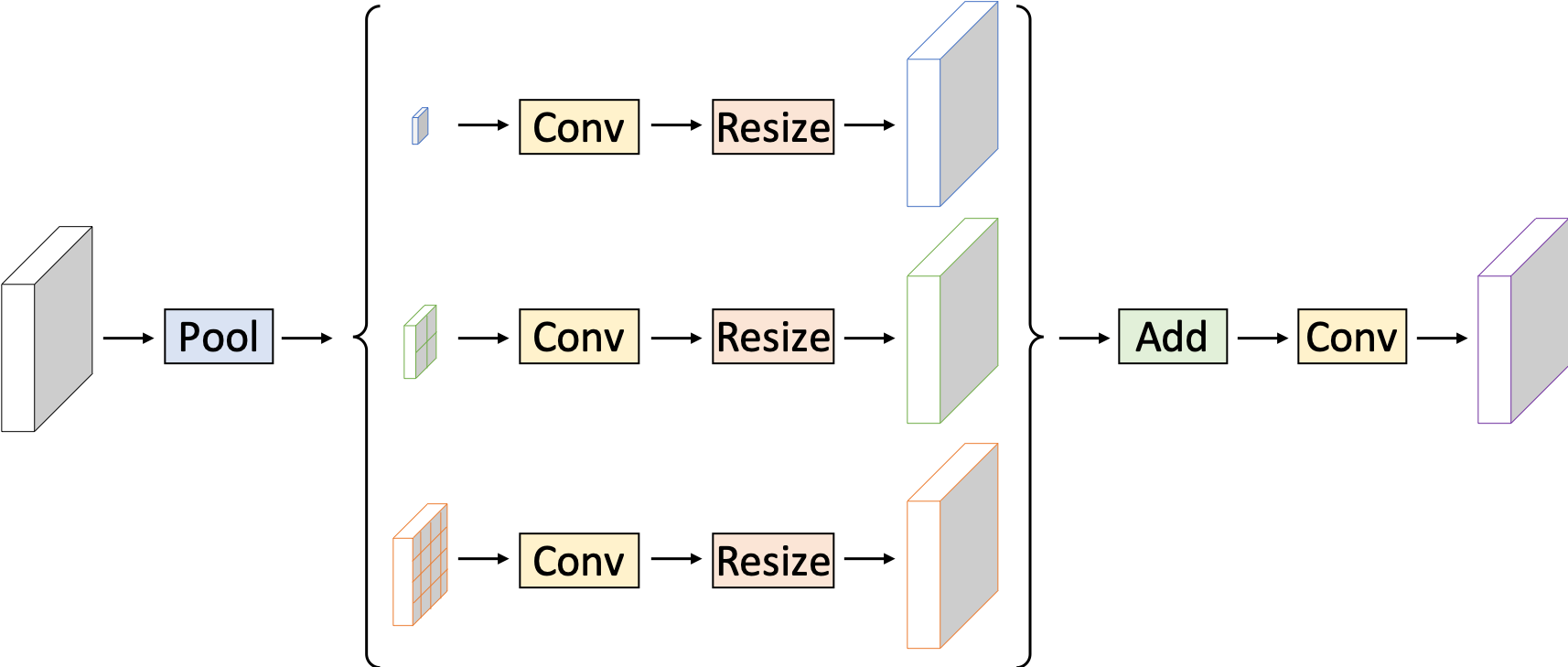}
   \caption{Simple Pyramid Pooling Module (SPPM). Conv denotes convolution, batch norm and relu operations. The bin sizes of global-average-pooling are $1\times1$, $2 \times 2$ and $4 \times 4$ respectively. }
   \label{fig:sppm}
\end{figure}

As shown in Figure \ref{fig:sppm}, we propose a Simple Pyramid Pooling Module (SPPM).
It first leverages the pyramid pooling module to fuse the input feature. The pyramid pooling module has three global-average-pooling operations and the bin sizes are $1\times1$, $2 \times 2$, and $4 \times 4$ respectively.
Afterwards, the output features are followed by the convolution and upsampling operations. For the convolution operation, the kernel size is $1 \times 1$ and the output channel is smaller than the input channel.
Finally, we add these upsampled features and apply a convolution operation to produce the refined feature.
Compared to original PPM, SPPM reduces the intermediate and output channels, removes the short-cut, and replaces the concatenate operation with an addition operation. Consequently, SPPM is more efficient and suitable for real-time models.

\subsection{Network Architecture}

The architecture of the proposed PP-LiteSeg is demonstrated in Figure \ref{fig:arch}. PP-LiteSeg mainly comprises three modules: encoder, aggregation and decoder.

Firstly, given an input image, PP-Lite utilizes a common lightweight network as encoder to extract hierarchical features.
In detail, we choose the STDCNet \cite{fan2021rethinking} for its outstanding performance.
The STDCNet has 5 stages, the stride for each stage is 2, so the final feature size is 1/32 of the input image. 
As shown in Table \ref{tab:ppliteseg_details}, we present two versions of PP-LiteSeg, i.e., PP-LiteSeg-T and PP-LiteSeg-B, of which the encoder is STDC1 and STDC2 respectively.
The PP-LiteSeg-B achieves higher segmentation accuracy, while the inference speed of PP-LiteSeg-T is faster.
It is worth noting that we apply the SSLD \cite{cui2021beyond} method to the training of the encoder and obtain enhanced pre-trained weights, which is beneficial for the convergence of segmentation training.

Secondly, PP-LiteSeg adopts SPPM to model the long-range dependencies. Taking the output feature of the encoder as input, SPPM produces a feature that contains global context information.

Finally, PP-LiteSeg utilizes our proposed FLD to gradually fuse multi-level features and output the resulting image.
Specifically, FLD consists of two UAFM and a segmentation head.
For efficiency, we adopt spatial attention module in UAFM.
Each UAFM takes two features as input, i.e., a low-level feature extracted by the stages of the encoder, a high-level feature generated by SPPM or the deeper fusion module.
The latter UAFM outputs fused features with the down-sample ratio of 1/8.
In the segmentation head, we perform Conv-BN-Relu operation to reduce the channels of the 1/8 down-sample feature to the number of classes.
An upsampling operation is followed to expand the feature size to the input image size, and an argmax operation predicts the label of each pixel.
The cross entropy loss with Online Hard Example Mining is adopted to optimize our models.

\begin{table}
  \centering
  \small
  \begin{tabular}{c | c c}
    \hline
    Model & Encoder & Channels in Decoder \\
    \hline
    PP-LiteSeg-T & STDC1 & 32, 64, 128 \\
    PP-LiteSeg-B & STDC2 & 64, 96, 128 \\
    \hline
  \end{tabular}
  \caption{The details of our proposed PP-LiteSeg.}
  \label{tab:ppliteseg_details}
\end{table}

\section{Experiments}
\label{sec:experiments}
In this section, we first introduce the datasets and implementation details.
Then, we compare the experimental results in terms of accuracy and inference speed with other state-of-the-art real-time methods.
Finally, we demonstrate the effectiveness of the proposed modules with the ablation study.

\subsection{Datasets and Implementation Details}

\noindent
\textbf{Cityscapes.}
The Cityscapes \cite{cordts2016cityscapes} is a large-scale dataset for urban segmentation.
It contains 5,000 fine annotated images, which are further split into 2975, 500, and 1525 images for training, validation and testing, respectively.
The resolution of the images is $2048 \times 1024$, which poses great challenges for the real-time semantic segmentation methods.
The annotated images have 30 classes and our experiments only use 19 classes for a fair comparison with other methods.

\noindent
\textbf{CamVid.}
Cambridge-driving Labeled Video Database (CamVid)  \cite{brostow2008segmentation} is a small-scale dataset for road scene segmentation.
There are 701 images with high-quality pixel-level annotations, in which 367, 101 and 233 images are chosen for training, validation and testing respectively. 
The images have the same resolution of $960  \times 720$.
The annotated images provide 32 categories, of which the subset of 11 categories are used in our experiments.

\noindent
\textbf{Training Settings.}
Following the common setting, the stochastic gradient descent (SGD) algorithm with 0.9 momentum is chosen as an optimizer.
We also adopt the warm-up strategy and the “poly” learning rate scheduler.
For Cityscapes, the batch size is 16, the max iterations are 160,000, the initial learning rate is 0.005, and the weight decay in the optimizer is $5e^{-4}$.
For CamVid, the batch size is 24, the max iterations is 1,000, the initial learning rate is 0.01, and the weight decay is $1e^{-4}$.
For data augmentation, we make use of random scaling, random cropping, random horizontal flipping, random color jittering and normalization. The random scale ranges in [0.125, 1.5], [0.5, 2.5] for Cityscapes and Camvid respectively. The cropped resolution of Cityscapes is $1024 \times 512$, and the cropped resolution of CamVid is $960 \times 720$.
All of our experiments are conducted on Tesla V100 GPU using PaddlePaddle\footnote{https://github.com/PaddlePaddle/Paddle}~\cite{ma2019paddlepaddle}. Code and pretrained models are available at PaddleSeg\footnote{https://github.com/PaddlePaddle/PaddleSeg}~\cite{liu2021paddleseg}.

\noindent
\textbf{Inference Settings.}
For a fair comparison, we export PP-LiteSeg to ONNX and utilize TensorRT to execute the model.
Similar to other methods \cite{yu2021bisenet, fan2021rethinking}, an image from Cityscapes is first resized to $1024 \times 512$ and $1536 \times 768$, then the inference model takes the scaled image and produces the predicted image, finally, the predicted image is resized to the original size of the input image.
The cost of the three steps is counted as the inference time.
For CamVid, the inference model takes the original image as input, while the resolution is $960 \times 720$.
We conduct all inference experiments under CUDA 10.2, CUDNN 7.6, TensorRT 7.1.3 on NVIDIA 1080Ti GPU.
We employ the standard mIoU for segmentation accuracy comparison and FPS for inference speed comparison.

\subsection{Experiments on Cityscapes}

\begin{figure*}[t]
  \centering
   \includegraphics[width=0.8\linewidth]{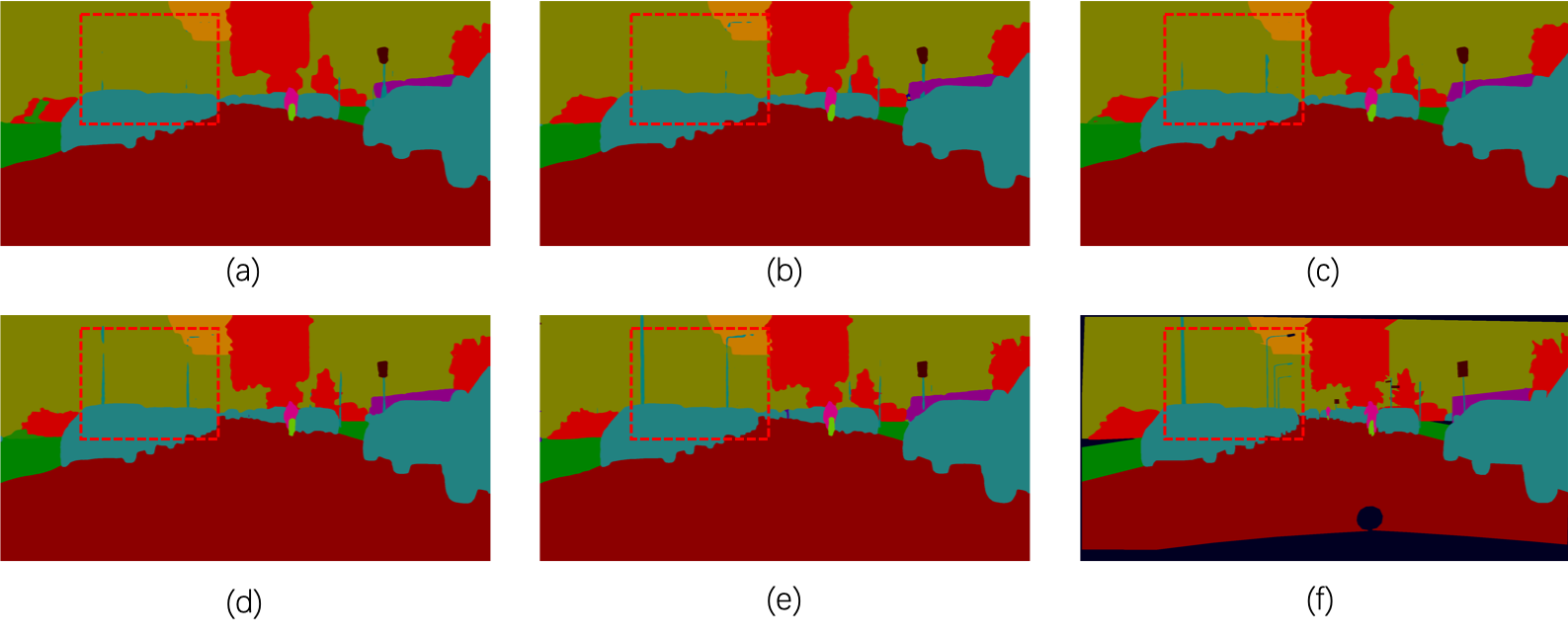}
   \caption{The qualitative comparison on the Cityscapes validation set. (a)-(e) represent the predicted image of baseline, baseline+FLD, baseline+FLD+SPPM, baseline+FLD+UAFM and baseline+FLD+SPPM+UAFM respectively, (f) denotes the ground truth.}
   \label{fig:ablation_proposed_models}
\end{figure*}

\subsubsection{Comparisons with State-of-the-arts}

\begin{table}
  \centering
  \small
  \resizebox{0.48 \textwidth}{!}{
      \begin{tabular}{l | c | c | c | c | c}
        \hline
        \multirow{2}{*}{Model} & \multirow{2}{*}{Encoder}  & \multirow{2}{*}{Resolution} & \multicolumn{2}{c|}{mIoU ($\%$)} & \multirow{2}{*}{FPS}  \\
        & & & val & test & \\
        \hline
        ENet \cite{paszke2016enet}         & -           &  $512 \times 1024$   & -    & 58.3 & 76.9  \\
        ICNet \cite{zhao2018icnet}       & PSPNet50    &  $1024 \times 2048$  & -    & 69.5 & 30.3  \\
        ESPNet \cite{mehta2018espnet}      & ESPNet      &  $512 \times 1024$   & -    & 60.3 & 112.9 \\
        ESPNetV2 \cite{mehta2019espnetv2}    & ESPNetV2    &  $512 \times 1024$   & 66.4 & 66.2 & -     \\
        SwiftNet \cite{orsic2019defense}     & ResNet18    &  $1024 \times 2048$  & 75.4 & 75.5 & 39.9  \\
        BiSeNetV1 \cite{yu2018bisenet}    & Xception39  &  $768 \times 1536$   & 69.0 & 68.4 & 105.8 \\
        BiSeNetV1-L \cite{yu2018bisenet}  & ResNet18    &  $768 \times 1536$   & 74.8 & 74.7 & 65.5  \\
        BiSeNetV2 \cite{yu2021bisenet}   & -           &  $512 \times 1024$   & 73.4 & 72.6 & 156   \\
        BiSeNetV2-L \cite{yu2021bisenet} & -           &  $512 \times 1024$   & 75.8 & 75.3 & 47.3  \\
        FasterSeg \cite{chen2019fasterseg}    & -           &  $1024 \times 2048$  & 73.1 & 71.5 & 163.9 \\
        SFNet \cite{li2020semantic}       & DF1         &  $1024 \times 2048$  & -    & 74.5 & 121   \\
        STDC1-Seg50 \cite{fan2021rethinking} & STDC1       &  $512 \times 1024$   & 72.2 & 71.9 & 250.4 \\
        STDC2-Seg50 \cite{fan2021rethinking} & STDC2       &  $512 \times 1024$   & 74.2 & 73.4 & 188.6 \\
        STDC1-Seg75 \cite{fan2021rethinking} & STDC1       &  $768 \times 1536$   & 74.5 & 75.3 & 126.7 \\
        STDC2-Seg75 \cite{fan2021rethinking} & STDC2       &  $768 \times 1536$   & 77.0 & 76.8 & 97.0 \\
        \hline
        PP-LiteSeg-T1 & STDC1 &  $512 \times 1024$  & 73.1 & 72.0 & \textbf{273.6}  \\
        PP-LiteSeg-B1 & STDC2 &  $512 \times 1024$  & 75.3 & 73.9 & 195.3 \\
        PP-LiteSeg-T2 & STDC1 &  $768 \times 1536$  & 76.0 & 74.9 & 143.6 \\
        PP-LiteSeg-B2 & STDC2 &  $768 \times 1536$  & \textbf{78.2} & \textbf{77.5} & 102.6\\
        \hline
      \end{tabular}
  }
  \caption{The comparisons with state-of-the-art real-time methods on Cityscapes. The training and inference setting refer to the implementation details.}
  \label{tab:acc_speed_cityscapes}
\end{table}

With the training and inference setting mentioned above, we compare the proposed PP-LiteSeg with previous state-of-the-art real-time models on Cityscapes.
For fair comparison, we evaluate PP-LiteSeg-T and PP-LiteSeg-B at two resolutions, i.e., $512 \times 1024$ and $768 \times 1536$.
Table \ref{tab:acc_speed_cityscapes} presents the model information, input resolution, mIoU, and FPS of various approaches.
Figure \ref{fig:speed_iou_fig} provides an intuitive comparison of segmentation accuracy and inference speed.
The experimental evaluations demonstrate that the proposed PP-LiteSeg achieves a state-of-the-art trade-off between accuracy and speed against other methods.
Specifically, we can observe PP-LiteSeg-T1 achieves 273.6 FPS and 72.0\% mIoU, which means the fastest inference speed and competitive accuracy.
With the resolution of $768 \times 1536$, PP-LiteSeg-B2 achieves the best accuracy, i,e. 78.2\% mIoU for the validation set, 77.5\% mIoU for the test set.
In addition, with same encoder and input resolution as STDC-Seg, PP-LiteSeg shows better performance.

\subsubsection{Ablation study}

\begin{table}
  \centering
  \small
  \begin{tabular}{c | c | c | c | c | c}
    \hline
    Model & FLD & SPPM & UAFM  & mIoU($\%$) & FPS \\
    \hline
    Baseline      &             &  &                        & 77.50  & 110.9 \\
    PP-LiteSeg-B2 & \checkmark  &  &                        & 77.67  & 109.7 \\
    PP-LiteSeg-B2 & \checkmark  & \checkmark  &             & 77.76  & 106.3 \\
    PP-LiteSeg-B2 & \checkmark  &  & \checkmark             & 77.89  & 105.5\\
    PP-LiteSeg-B2 & \checkmark  & \checkmark  & \checkmark  & 78.21  & 102.6 \\
    \hline
  \end{tabular}
  \caption{Ablation study for our proposed modules on the validation set of Cityscapes. The baseline model is PP-LiteSeg-B2 without the proposed modules.}
  \label{tab:ablation}
\end{table}

Ablation experiments are conducted to demonstrate the effectiveness of the proposed modules.
The experiments choose PP-LiteSeg-B2 in the comparison and use the same training and inference setting.
The baseline model is the PP-LiteSeg-B2 without the proposed modules, while the number of features channels is 96 in decoder and the fusion method is element-wise summation. 
Table \ref{tab:ablation} presents the quantitative results of our ablation study.
We can find that the FLD in PP-LiteSeg-B2 improves the mIoU by 0.17\%.
Adding SPPM and UAFM also improve the segmentation accuracy, while the inference speed slightly decreases.
Based on three proposed modules, PP-LiteSeg-B2 achieves 78.21 mIoU with 102.6 FPS. The mIoU is boosted by 0.71\% compared to the baseline model.
Figure \ref{fig:ablation_proposed_models} provides the qualitative comparisons.
We can observe that the predicted image becomes more consistent with the ground truth when adding FLD, SPPM and UAFM one by one.
In short, our proposed modules are effective for semantic segmentation.

\subsection{Experiments on CamVid}

\begin{table}
  \centering
  \small
  \resizebox{0.48 \textwidth}{!}{
      \begin{tabular}{l | c | c | c }
        \hline
        Model & Encoder  & mIoU ($\%$) & FPS  \\
        \hline
        ENet \cite{paszke2016enet}       & -             & 51.3 &  61.2 \\
        ICNet \cite{zhao2018icnet}       & PSPNet50      & 67.1 &  34.5  \\
        DFANet A  \cite{li2019dfanet}    & Xception A    & 64.7 &  120  \\
        SwiftNet \cite{orsic2019defense} & ResNet18      & 72.58 & -  \\
        BiSeNetV1 \cite{yu2018bisenet}   & Xception39    & 65.6 & 175  \\
        BiSeNetV1-L \cite{yu2018bisenet} & ResNet18      & 68.7 & 116.3   \\
        BiSeNetV2 \cite{yu2021bisenet}   & -             & 72.4 & 124.5  \\
        BiSeNetV2-L \cite{yu2021bisenet} & -             & 73.2 & 32.7   \\
        STDC1-Seg \cite{fan2021rethinking} & STDC1       & 73.0 & 197.6  \\
        STDC2-Seg \cite{fan2021rethinking} & STDC2       & 73.9 & 152.2  \\
        \hline
        PP-LiteSeg-T & STDC1   & 73.3 & \textbf{222.3} \\
        PP-LiteSeg-B & STDC2  & \textbf{75.0} & 154.8 \\
        \hline
      \end{tabular}
  }
  \caption{The comparisons with state-of-the-art real-time methods on CamVid test set. The input resolution of all methods is $960 \times 720$.}
  \label{tab:acc_speed_camvid}
\end{table}

To further demonstrate the capability of PP-LiteSeg, we also conduct experiments on the CamVid dataset.
Similar to other works, the input resolution for training and inference is $960 \times 720$.
As shown in Table \ref{tab:acc_speed_camvid}, PP-LiteSeg-T achieves 222.3 FPS, which is over 12.5\% faster than other methods.
PP-LiteSeg-B achieves the best accuracy, i.e., 75.0\% mIoU with 154.8 FPS.
Overall, the comparisons show PP-LiteSeg achieves a state-of-the-art trade-off between accuracy and speed on Camvid.

\section{Conclusions}
\label{sec:conclusions}

In this paper, we focus on designing a novel real-time network for semantic segmentation.
First, Flexible and Lightweight Decoder (FLD) is proposed to improve the efficiency of previous decoder.
Then, we present a Unified Attention Fusion Module (UAFM), which is effective for strengthening feature representations.
Furthermore, we propose a Simple Pyramid Pooling Module (SPPM) to aggregate global context with low computation cost.
Based on these novel modules, we propose PP-LiteSeg, a real-time semantic segmentation network.
Extensive experiments demonstrate that PP-LiteSeg achieves a state-of-the-art trade-off between segmentation accuracy and inference speed.
In future work, we plan to apply our methods into more tasks, such as matting and interactive segmentation.

{\small
\bibliographystyle{ieee_fullname}
\bibliography{egbib}

\begin{thebibliography}{10}\itemsep=-1pt

\bibitem{badrinarayanan2017segnet}
Vijay Badrinarayanan, Alex Kendall, and Roberto Cipolla.
\newblock Segnet: A deep convolutional encoder-decoder architecture for image
  segmentation.
\newblock {\em IEEE transactions on pattern analysis and machine intelligence},
  39(12):2481--2495, 2017.

\bibitem{brostow2008segmentation}
Gabriel~J Brostow, Jamie Shotton, Julien Fauqueur, and Roberto Cipolla.
\newblock Segmentation and recognition using structure from motion point
  clouds.
\newblock In {\em European conference on computer vision}, pages 44--57.
  Springer, 2008.

\bibitem{chen2018encoder}
Liang-Chieh Chen, Yukun Zhu, George Papandreou, Florian Schroff, and Hartwig
  Adam.
\newblock Encoder-decoder with atrous separable convolution for semantic image
  segmentation.
\newblock In {\em Proceedings of the European conference on computer vision
  (ECCV)}, pages 801--818, 2018.

\bibitem{chen2019fasterseg}
Wuyang Chen, Xinyu Gong, Xianming Liu, Qian Zhang, Yuan Li, and Zhangyang Wang.
\newblock Fasterseg: Searching for faster real-time semantic segmentation.
\newblock {\em arXiv preprint arXiv:1912.10917}, 2019.

\bibitem{chu2022pp}
Lutao Chu, Yi Liu, Zewu Wu, Shiyu Tang, Guowei Chen, Yuying Hao, Juncai Peng,
  Zhiliang Yu, Zeyu Chen, Baohua Lai, et~al.
\newblock Pp-humanseg: Connectivity-aware portrait segmentation with a
  large-scale teleconferencing video dataset.
\newblock In {\em Proceedings of the IEEE/CVF Winter Conference on Applications
  of Computer Vision}, pages 202--209, 2022.

\bibitem{cordts2016cityscapes}
Marius Cordts, Mohamed Omran, Sebastian Ramos, Timo Rehfeld, Markus Enzweiler,
  Rodrigo Benenson, Uwe Franke, Stefan Roth, and Bernt Schiele.
\newblock The cityscapes dataset for semantic urban scene understanding.
\newblock In {\em Proceedings of the IEEE conference on computer vision and
  pattern recognition}, pages 3213--3223, 2016.

\bibitem{cui2021beyond}
Cheng Cui, Ruoyu Guo, Yuning Du, Dongliang He, Fu Li, Zewu Wu, Qiwen Liu,
  Shilei Wen, Jizhou Huang, Xiaoguang Hu, et~al.
\newblock Beyond self-supervision: A simple yet effective network distillation
  alternative to improve backbones.
\newblock {\em arXiv preprint arXiv:2103.05959}, 2021.

\bibitem{fan2021rethinking}
Mingyuan Fan, Shenqi Lai, Junshi Huang, Xiaoming Wei, Zhenhua Chai, Junfeng
  Luo, and Xiaolin Wei.
\newblock Rethinking bisenet for real-time semantic segmentation.
\newblock In {\em Proceedings of the IEEE/CVF conference on computer vision and
  pattern recognition}, pages 9716--9725, 2021.

\bibitem{hao2021edgeflow}
Yuying Hao, Yi Liu, Zewu Wu, Lin Han, Yizhou Chen, Guowei Chen, Lutao Chu,
  Shiyu Tang, Zhiliang Yu, Zeyu Chen, et~al.
\newblock Edgeflow: Achieving practical interactive segmentation with
  edge-guided flow.
\newblock In {\em Proceedings of the IEEE/CVF International Conference on
  Computer Vision}, pages 1551--1560, 2021.

\bibitem{hou2019learning}
Yuenan Hou, Zheng Ma, Chunxiao Liu, and Chen~Change Loy.
\newblock Learning lightweight lane detection cnns by self attention
  distillation.
\newblock In {\em Proceedings of the IEEE/CVF international conference on
  computer vision}, pages 1013--1021, 2019.

\bibitem{huang2021fapn}
Shihua Huang, Zhichao Lu, Ran Cheng, and Cheng He.
\newblock Fapn: Feature-aligned pyramid network for dense image prediction.
\newblock In {\em Proceedings of the IEEE/CVF International Conference on
  Computer Vision}, pages 864--873, 2021.

\bibitem{huang2021alignseg}
Zilong Huang, Yunchao Wei, Xinggang Wang, Wenyu Liu, Thomas~S Huang, and
  Humphrey Shi.
\newblock Alignseg: Feature-aligned segmentation networks.
\newblock {\em IEEE Transactions on Pattern Analysis and Machine Intelligence},
  44(1):550--557, 2021.

\bibitem{lateef2019survey}
Fahad Lateef and Yassine Ruichek.
\newblock Survey on semantic segmentation using deep learning techniques.
\newblock {\em Neurocomputing}, 338:321--348, 2019.

\bibitem{li2019dfanet}
Hanchao Li, Pengfei Xiong, Haoqiang Fan, and Jian Sun.
\newblock Dfanet: Deep feature aggregation for real-time semantic segmentation.
\newblock In {\em Proceedings of the IEEE/CVF Conference on Computer Vision and
  Pattern Recognition}, pages 9522--9531, 2019.

\bibitem{li2020semantic}
Xiangtai Li, Ansheng You, Zhen Zhu, Houlong Zhao, Maoke Yang, Kuiyuan Yang,
  Shaohua Tan, and Yunhai Tong.
\newblock Semantic flow for fast and accurate scene parsing.
\newblock In {\em European Conference on Computer Vision}, pages 775--793.
  Springer, 2020.

\bibitem{liu2021paddleseg}
Yi Liu, Lutao Chu, Guowei Chen, Zewu Wu, Zeyu Chen, Baohua Lai, and Yuying Hao.
\newblock Paddleseg: A high-efficient development toolkit for image
  segmentation.
\newblock {\em arXiv preprint arXiv:2101.06175}, 2021.

\bibitem{liu2021swin}
Ze Liu, Han Hu, Yutong Lin, Zhuliang Yao, Zhenda Xie, Yixuan Wei, Jia Ning, Yue
  Cao, Zheng Zhang, Li Dong, et~al.
\newblock Swin transformer v2: Scaling up capacity and resolution.
\newblock {\em arXiv preprint arXiv:2111.09883}, 2021.

\bibitem{long2015fully}
Jonathan Long, Evan Shelhamer, and Trevor Darrell.
\newblock Fully convolutional networks for semantic segmentation.
\newblock In {\em Proceedings of the IEEE conference on computer vision and
  pattern recognition}, pages 3431--3440, 2015.

\bibitem{ma2019paddlepaddle}
Yanjun Ma, Dianhai Yu, Tian Wu, and Haifeng Wang.
\newblock Paddlepaddle: An open-source deep learning platform from industrial
  practice.
\newblock {\em Frontiers of Data and Domputing}, 1(1):105--115, 2019.

\bibitem{mehta2018espnet}
Sachin Mehta, Mohammad Rastegari, Anat Caspi, Linda Shapiro, and Hannaneh
  Hajishirzi.
\newblock Espnet: Efficient spatial pyramid of dilated convolutions for
  semantic segmentation.
\newblock In {\em Proceedings of the european conference on computer vision
  (ECCV)}, pages 552--568, 2018.

\bibitem{mehta2019espnetv2}
Sachin Mehta, Mohammad Rastegari, Linda Shapiro, and Hannaneh Hajishirzi.
\newblock Espnetv2: A light-weight, power efficient, and general purpose
  convolutional neural network.
\newblock In {\em Proceedings of the IEEE/CVF conference on computer vision and
  pattern recognition}, pages 9190--9200, 2019.

\bibitem{orsic2019defense}
Marin Orsic, Ivan Kreso, Petra Bevandic, and Sinisa Segvic.
\newblock In defense of pre-trained imagenet architectures for real-time
  semantic segmentation of road-driving images.
\newblock In {\em Proceedings of the IEEE/CVF Conference on Computer Vision and
  Pattern Recognition}, pages 12607--12616, 2019.

\bibitem{paszke2016enet}
Adam Paszke, Abhishek Chaurasia, Sangpil Kim, and Eugenio Culurciello.
\newblock Enet: A deep neural network architecture for real-time semantic
  segmentation.
\newblock {\em arXiv preprint arXiv:1606.02147}, 2016.

\bibitem{siam2018comparative}
Mennatullah Siam, Mostafa Gamal, Moemen Abdel-Razek, Senthil Yogamani, Martin
  Jagersand, and Hong Zhang.
\newblock A comparative study of real-time semantic segmentation for autonomous
  driving.
\newblock In {\em Proceedings of the IEEE conference on computer vision and
  pattern recognition workshops}, pages 587--597, 2018.

\bibitem{song2021attanet}
Qi Song, Kangfu Mei, and Rui Huang.
\newblock Attanet: Attention-augmented network for fast and accurate scene
  parsing.
\newblock In {\em AAAI}, 2021.

\bibitem{yu2021bisenet}
Changqian Yu, Changxin Gao, Jingbo Wang, Gang Yu, Chunhua Shen, and Nong Sang.
\newblock Bisenet v2: Bilateral network with guided aggregation for real-time
  semantic segmentation.
\newblock {\em International Journal of Computer Vision}, 129(11):3051--3068,
  2021.

\bibitem{yu2018bisenet}
Changqian Yu, Jingbo Wang, Chao Peng, Changxin Gao, Gang Yu, and Nong Sang.
\newblock Bisenet: Bilateral segmentation network for real-time semantic
  segmentation.
\newblock In {\em Proceedings of the European conference on computer vision
  (ECCV)}, pages 325--341, 2018.

\bibitem{zhao2018icnet}
Hengshuang Zhao, Xiaojuan Qi, Xiaoyong Shen, Jianping Shi, and Jiaya Jia.
\newblock Icnet for real-time semantic segmentation on high-resolution images.
\newblock In {\em Proceedings of the European conference on computer vision
  (ECCV)}, pages 405--420, 2018.

\bibitem{zhao2017pyramid}
Hengshuang Zhao, Jianping Shi, Xiaojuan Qi, Xiaogang Wang, and Jiaya Jia.
\newblock Pyramid scene parsing network.
\newblock In {\em Proceedings of the IEEE conference on computer vision and
  pattern recognition}, pages 2881--2890, 2017.

\bibitem{zhou2018unet++}
Zongwei Zhou, Md~Mahfuzur Rahman~Siddiquee, Nima Tajbakhsh, and Jianming Liang.
\newblock Unet++: A nested u-net architecture for medical image segmentation.
\newblock In {\em Deep learning in medical image analysis and multimodal
  learning for clinical decision support}, pages 3--11. Springer, 2018.

\end{thebibliography}
}

\end{document}